# ROBIN: A Robust Optical Binary Neural Network Accelerator

FEBIN P. SUNNY, ASIF MIRZA, MAHDI NIKDAST, SUDEEP PASRICHA

Department of Electrical and Computer Engineering, Colorado State University

Domain specific neural network accelerators have garnered attention because of their improved energy efficiency and inference performance compared to CPUs and GPUs. Such accelerators are thus well suited for resource-constrained embedded systems. However, mapping sophisticated neural network models on these accelerators still entails significant energy and memory consumption, along with high inference time overhead. Binarized neural networks (BNNs), which utilize single-bit weights, represent an efficient way to implement and deploy neural network models on accelerators. In this paper, we present a novel optical-domain BNN accelerator, named *ROBIN*, which intelligently integrates heterogeneous microring resonator optical devices with complementary capabilities to efficiently implement the key functionalities in BNNs. We perform detailed fabrication-process variation analyses at the optical device level, explore efficient corrective tuning for these devices, and integrate circuit-level optimization to counter thermal variations. As a result, our proposed *ROBIN* architecture possesses the desirable traits of being robust, energy-efficient, low latency, and high throughput, when executing BNN models. Our analysis shows that *ROBIN* can outperform the best-known optical BNN accelerators and also many electronic accelerators. Specifically, our energy-efficient *ROBIN* design exhibits energy-per-bit values that are ~4× lower than electronic BNN accelerators and ~933× lower than a recently proposed photonic BNN accelerator, while a performance-efficient *ROBIN* design shows ~3x and ~25x better performance than electronic and photonic BNN accelerators, respectively.

## 1 INTRODUCTION

Machine learning applications have become increasingly prevalent in the last decade with many emerging applications, such as autonomous transportation, medical prognosis, real-time speech translation, network anomaly detection, and audio/video synthesis. This prevalence is fueled by the emergence of sophisticated and powerful machine learning models over the past decade, such as Deep Neural Networks (DNNs) and Convolutional Neural Networks (CNNs). As researchers explore deeper models with higher connectivity, the compute power and the memory requirement necessary to train and deploy them also increase. Such increasing complexity also necessitates that the underlying hardware platforms consistently deliver better performance while satisfying strict power requirements. This endeavor to achieve high performance-per-watt has driven hardware architects to design custom accelerators for deep learning, e.g., Google's TPU [1] and Intel's Movidius [2], with much higher performance-per-watt than CPUs and GPUs. The performance-per-watt requirement still remains a challenge in resource-constrained environments such as many embedded devices, where computational power, energy expenditure, and available memory are often limited. Binarized Neural Networks (BNNs) [3], [4] are a possible solution to this challenge, as they reduce memory and computational requirements of DNN and CNN models while offering competitive accuracies with full precision models, when executed on custom accelerators.

Another potential solution to reduce-performance-per-watt for neural-network processing is to explore more efficient hardware accelerator platforms. Conventional electronic accelerator platforms face fundamental limits in the post-Moore era where the high costs and diminishing performance improvements with semiconductor-technology scaling prevent significant improvements in future product generations [5]. A well-known bottleneck in these accelerators is moving data electronically on metallic wires, which puts limits on achievable performance (e.g., bandwidth and latency) and energy savings [6]. A solution to the data-movement bottleneck has presented itself in the form of silicon photonics technology,

This research is supported by grants from NSF (CCF-1813370, CCF-2006788).

.Authors' addresses: F. P. Sunny, A. Mirza, M. Nikdast, and S. Pasricha (corresponding author), Department of Electrical and Computer Engineering, Colorado State University, Fort Collins, CO 80523-1373; email: { febin.sunny, mirza.baig, mahdi.nikdast, sudeep}@colostate.edu.

which enables ultra-high bandwidth, low-latency, and energy-efficient communication [7]. CMOS-compatible optical interconnects have already replaced metallic ones for light-speed data transmission at almost every level of computing, and are now actively being considered for chip-scale integration [8]. Interestingly, it is also possible to use optical components to perform computation, e.g., matrix-vector multiplication [9]. Due to the emergence of both chip-scale optical communication and computation, it is now possible to conceive photonic integrated circuits (PICs) that offer low latency and energy-efficient optical domain data transport and computation.

Despite the benefits of utilizing photonics for computation and communication, there are some obstacles that must be addressed before photonic accelerators become truly viable. One of the main obstacles that impacts the robustness and reliability of photonic accelerators is their sensitivity to fabrication process and thermal variations. These variations introduce undesirable crosstalk, optical phase shifts, frequency drifts, tuning overheads, and photodetection current mismatches, which adversely affect the reliable and robust operation of photonic accelerators. In order to correct the impact of variations, thermo-optic (TO) or electro-optic (EO) tuning circuits are often used, which have notable power overheads. Because of the phase-change effects it has on photonic devices, TO tuning may also be used to control weight/activation imprinting via microring resonators (MRs). But the high latency of operation (in $\mu s$ range [11]) of TO tuning can limit the achievable throughput and parallelism in photonic accelerators.

In this paper, we introduce *ROBIN*, a novel optical-domain BNN accelerator that addresses the challenges highlighted above by optimizing electro-optic components across the device, circuit, and architecture layers. *ROBIN* combines novel device- and circuit-level techniques to achieve more efficient fabrication-process-variation (FPV) correction in optical devices, which helps with reducing energy and improving accuracy in BNNs that utilize these devices. Additionally, circuit-level tuning enhancements for inference latency reduction, and an optimized architecture-level design help improve performance and also energy consumption compared to the state-of-the-art. Our novel contributions in this work include:

- The design of a novel optical-domain BNN accelerator architecture that is robust to fabrication process variations (FPVs) and thermal variations, and utilizes efficient wavelength reuse and a modular structure to enable high throughput and energy-efficient execution across BNN models;
- A novel integration of heterogeneous optical microring resonator (MR) devices; we also conduct design space exploration for these MR designs to determine device characteristics for efficient BNN execution;
- An enhanced tuning circuit to simultaneously support large thermal-induced resonance shifts and high-speed, low-loss device tuning to compensate for FPVs;
- A comprehensive comparison with state-of-the-art BNN and non-BNN accelerator platforms from the optical and electronic domains, to demonstrate the potential of our BNN accelerator platform.

The rest of the paper is organized as follows: Section 2 briefly explores the related works in the field of BNN acceleration. Section 3 gives a brief overview of non-coherent optical computation for photonic accelerators similar to ours. Section 4 provides an overview of BNNs and the partially binarized approach we have adopted for better accuracy in models. Section 5 describes the ROBIN architecture and our optimization efforts in tuning circuits, photonic devices, and photonic system level. Details of the experiments conducted, simulation setup, and the obtained results are provided in Section 6. Finally, Section 7 presents some concluding remarks and directions for future work.

## 2 RELATED WORK

Silicon-photonic-based DNN accelerator architectures are becoming increasingly prominent with significant interest from both academic and industrial research communities [12]. This growth in interest can be attributed to the benefits of photonic acceleration over electronic acceleration, as discussed in the previous section. Optical DNN accelerator architectures can be broadly classified into two types: coherent architectures and non-coherent architectures. Coherent architectures use a single wavelength to operate and imprint weight/activation parameters onto the electrical field amplitude of the light wave [10], [13]. These architectures mainly use on-chip optical interferometer devices like Mach-Zehnder Interferometers (MZIs). For imprinting the parameters, optical phase-change mechanisms can be introduced to MZI devices. These mechanisms use heating or carrier injection to change the refractive index in the MZI structure. Weighting occurs with electrical field amplitude attenuation proportional to the weight value, and phase modulation that is proportional to the sign of the weight. The weighted signals are then accumulated with cascaded optical combiners, through coherent interference. Here the term coherent refers to the physical property of the wave, where it is possible for the wave to interfere constructively or destructively, on the same wavelength. Non-coherent architectures, such as [14]-[16], use multiple wavelengths, where each wavelength can be used to perform an individual neuron operation. These architectures are referred to as non-coherent architectures as they use different optical wavelengths, the interaction among which can be non-coherent. A large number of neurons can be represented simultaneously in non-coherent architectures by using wavelength-division multiplexing (WDM) or dense WDM (DWDM). In these architectures, parameters are imprinted on to the signal amplitude directly, and to manipulate individual wavelengths, wavelength-selective devices such as microring resonators (MRs) or microdisks are used. The optical signal power is controlled, for imprinting parameter values, by controlling the optical loss in these devices through tuning mechanisms (Section 5.1). The Broadcast and Weight (B&W) protocol [17] is typically employed for setting and updating the weight and activation values. The *ROBIN* architecture we present in this paper is a noncoherent architecture, i.e., it uses multiple wavelengths that are routed to photonic computation units in waveguides using WDM in accordance with the B&W protocol. The growing interest in noncoherent architectures can be attributed to the limitations in scalability, phase encoding noise, and phase error accumulation in coherent architectures [12], [18].

    For optical DNN acceleration using noncoherent mechanics, [14] introduced a photonic accelerator for CNNs where all the layers of CNN models are implemented using connected photonic convolution units. In these units, MRs are used to tune wavelengths to desired kernel values through phase tuning. Another such work, in [15], utilizes microdisks instead of MRs due to the lower area and power consumption they offer. But microdisks use 'whispering gallery mode' resonance which is inherently lossy due to the tunneling ray attenuation phenomenon [19]; which reduces reliability and energy-efficiency with microdisks. There are very few works which focus on implementations of BNN accelerators with optical components. The work in [20] proposed an MR-based accelerator for discretized neural network acceleration, with an encoding scheme to enable positive and negative product considerations. The authors in [21] leveraged microdisks for implementing an accelerator with a design similar to [15]. This work considered an accelerator for fully binarized neural networks, i.e., both weights and activations and considered to be single-bit parameters. Because of this simplification, [21] was able to utilize energy-efficient photonic XOR and population count operations instead of conventional multiply and accumulate operations. The work also made use of photonic non-volatile memory and claimed operating frequencies of up to 50 GHz. All of these existing works on noncoherent optical-domain DNN/BNN acceleration have several shortcomings. They suffer from susceptibility to fabrications process variations (FPVs) and also thermal crosstalk, which are not addressed in these architectures. Microsecond-granularity thermo-optic tuning latencies further can reduce the speed and efficiency of optical computing [11], which is also not considered when analyzing accelerator performance. We address these crucial shortcomings as part of our *ROBIN* optical-domain BNN accelerator architecture in this work.

    In this work, we aim to ensure the robustness of the architecture against process and thermal variations by using MR design-space exploration and photonic tuning-circuit optimizations, which will be further explained in Section 5. We also utilize the broadband capabilities of the key photonic device in our work, microring resonators (MRs), to perform batch normalization folding, which moves batch normalization operations from the electrical domain to the photonic domain.

Section 5.3 further details the modular architectural design aiming at ensuring wavelength reuse, to reduce VCSEL usage and splitter losses and waveguide length reduction. We also explore how the architecture performs in the presence of FPVs and how we may further reduce energy consumption in terms of device tuning in this scenario, in Section 6.2.

## 3 OVERVIEW OF NONCOHERENT OPTICAL COMPUTATION

Noncoherent optical accelerators leverage the low-latency and energy-efficient optical computation for multiply and accumulate (MAC) operations, which consumes substantial computational power and incurs high latencies in electronic accelerators. These accelerators typically utilize the B&W protocol with multiple wavelengths. Figure 1(a) (from [22]) gives an overview of a B&W-based optical MAC unit. The figure depicts a recurrent MAC unit which is employed repeatedly to compute different layers of a neural network model. The layer parameters such as weights or activations can be imprinted on to the wavelengths using the MRs that are tuned to modify the optical signal intensity to represent those values. The MRs are placed in MR banks where multiple parameters can be imprinted onto wavelengths simultaneously. In the MR banks, each MR is tuned to a specific optical wavelength and can be used to alter the optical characteristics of the wavelength (thereby changing its intensity) to represent the imprinted parameter. There can be separate wavelengths which carry positive and negative parameters, as discussed in [20]; these parameters are summed using balanced photodetectors (BPDs), as shown in Figure 1(a).

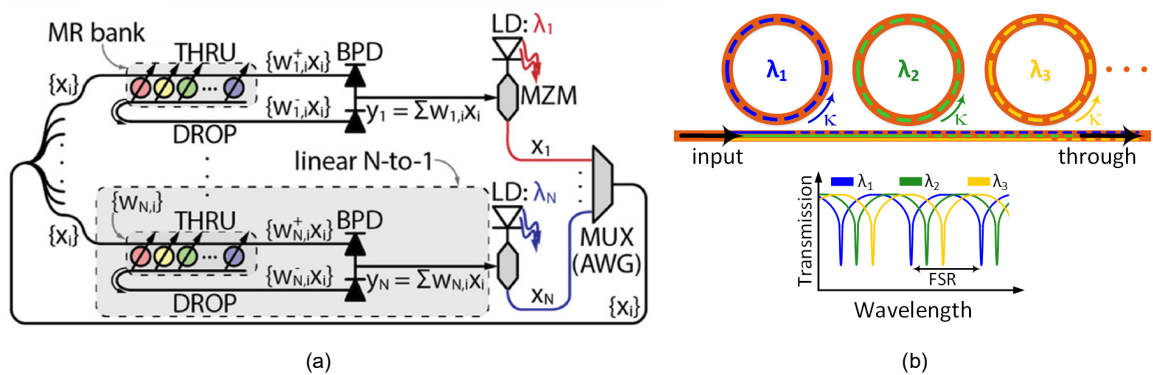

Figure 1: (a) A recurrent noncoherent B&W MAC based design [22]; (b) An MR bank consisting of MRs with individual resonant wavelength ($\lambda_i$) coupled to the MRs at cross-over coupling ($\kappa$) and the output spectrum, showing free spectral range (FSR).

The output from the MAC unit is passed on to a Mach-Zehnder Modulator (MZM) which tunes the output from a designated laser diode (LD) to this output. Multiple MZMs and LDs are used to generate the outputs from multiple MAC units; these are collected and multiplexed using an arrayed waveguide grating (AWG) based optical multiplexer (MUX). The output from the MUX is passed back into the MAC units, through splitters, now embedded with parameters from the next layer. Devices such as electro-optic modulators (not depicted) may be used to implement non-linearities after the MAC operation. Unfortunately, the static nature of the hardware limits the size of the neural network model that can be accelerated using such a configuration. The number of splitters being used can also cause increased optical losses and thus higher laser power requirement to compensate for the losses as the size of an accelerator using this B&W configuration increases.

MRs and other on-chip optical resonators such as microdisks are crucial components in such noncoherent MAC configurations, as they impact the reliability and efficiency of the operation performed. Figure 1(b) depicts an MR bank and its output spectrum along with the free spectral range (FSR). Factors such as fabrication-process variations (FPVs) and thermal variations which impact the MR critical dimensions and hence the effective refractive index ($n_{eff}$) of the device can cause a drift in the resonant wavelength ($\Delta\lambda_{MR}$) [23]. This drift can introduce errors into optical computation and is thus usually compensated for (i.e., corrected) with TO or EO tuning circuits. While EO offers faster tuning (~ns range) and consumes lesser power (~4 $\mu m$/nm), it also has a smaller tuning range [24]. TO tuning, on the other hand, consumes higher power (~27 mW/FSR) and has higher tuning latency (~$\mu s$ range) [11], but offers a larger tuning range. Because of the larger correction capacity, TO is often preferred over EO despite its higher latency and power consumption. Therefore, as the number of MRs increases—when considering larger CNN or MLP models—the tuning power consumption also increases. This also creates increased wavelength requirements per waveguide and calls for longer waveguides to host the MRs, causing increased laser power consumption to supply the wavelengths and to compensate for the propagation losses in the longer waveguides. Also, more MRs and more wavelengths increase optical crosstalk, and also introduce thermal crosstalk due to the larger number of TO tuners employed. To counteract these challenges, and ensure better weight resolution, crosstalk mitigation strategies must also be considered.

To design an effective optical-domain BNN accelerator, all of these considerations must be taken into account. This in turn highlights the need for *(i)* better device optimizations to tolerate variations; *(ii)* efficient and low-latency tuning mechanisms; *(iii)* a scalable architecture design, which is optimized for energy efficiency, area, and throughput. Our work aims to address all of these concerns for an efficient BNN accelerator implementation in the optical domain.

## 4 BINARIZED NEURAL NETWORKS

BNNs [3] are types of DNNs (or CNNs) where both weights and activation parameters only use binary values, and the binary values are utilized during both inference and backpropagation training. The binary nature of weights in BNNs makes them resilient to small perturbations which can usually lead to gross classification errors in DNNs. Inspired by the seminal work on efficiently training BNNs [3], recent efforts either explore how BNN accuracy can be improved, apply BNNs to different application domains, or explore how BNNs can be implemented efficiently in hardware to leverage their low computation power and memory requirements in resource constrained environments. .

BNNs utilize the sign function to convert real valued weights to +1 or -1. But this typically leads to complications in training as the gradient for the sign function always results in a zero. The work in [25] introduced a heuristic called straight through estimator (STE) to circumvent this issue. STEs approximate the gradient by bypassing the gradient of the layer, by turning it into an identity function. The gradient thus obtained is used for updating real valued weights, using an optimization strategy such as Adam or stochastic gradient descent (SGD). This process is utilized for activation parameters as well. Also, the use of batch normalization (BN) layers in BNNs has been shown to lead to several benefits [4]. The gain ($\gamma$) and bias ($\beta$) terms of the BN layer not only help condition the values during training, which speeds up BNN training, but also helps to improve accuracy in BNNs.

Another approach for increasing inference accuracy in BNNs is to employ partial binarization, where some of the layers are not completely binarized. The last layer is usually not binarized to avoid severe loss in accuracy. With detailed analysis of the model, critical layers can be identified and can be kept at higher precision, for better accuracy, at the cost of increased resource (computation, memory) utilization. To determine the appropriate activation parameter precision, which is required to determine the digital-to-analog converter (DAC) resolution in our accelerator architecture, we conducted a BNN accuracy analysis, where weights were restricted to binary (1-bit) values, but the bit precision level of the activations was altered from 1-bit to 16-bits. During BNN training, we ensured that we only binarize weights during the forward and backward propagations but not during the parameter update step, because keeping good precision weights during the updates is necessary for SGD to work at all (as parameter changes are usually tiny during gradient descent). After training, all weights were in binary format, while the precision of input activations was varied. Figure 2 shows the results of varying activation precision across four different models and their datasets (described later in Section 6.1). We observed that the accuracy had notable change initially as activations bits were increased, but this gain in accuracy soon saturated. Based on the results, we consider 1-bit weights with 4-bit activations, and thus use 4-bit DACs in our architecture.

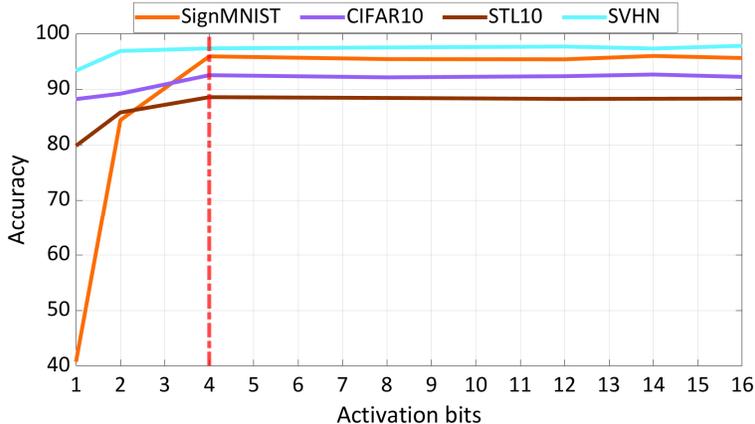

Figure 2: The accuracy sensitivity study conducted by varying activation parameter precision (number of bits). Weights are kept as binary values in all cases. The study was performed across four different models and their datasets (described later in Section 6.1).

## 5 ROBIN ARCHITECTURE

In this section, we describe the various optimization considerations at device, circuit, and architecture level used for designing the *ROBIN* architecture.

### 5.1 Tuning circuit design

A tuning circuit design is essential for fast and accurate operation of MRs in our BNN accelerator. Errors due to fabrication process variations (FPVs) can be significantly reduced by using an appropriate MR tuning circuit. The tuning circuit employed can be either thermo-optic (TO) or electro-optic (EO) tuning circuits. Thermo-optic(TO)-based tuning mechanisms use microheaters to change the temperature in the proximity of a microring resonator (MR), which then alters the effective index ($n_{eff}$) of the MR. This in turn changes the device characteristics such as resonant wavelength ($\lambda_{MR}$). Such a change in resonant wavelength ($\Delta\lambda_{MR}$) can help compensate for fabrication-process and thermal variations in MRs. The electro-optic (EO)-based tuning mechanisms in an MR is based on the depletion and injection of carriers on a PN diode. EO tuning is faster ($\sim ns$ range) and consumes lower power ($4\ \mu W/nm$) [26] when compared to TO tuning ($27\ mW/FSR$) [27], where FSR is the free-spectral range. However, only small shifts in an MR's resonant wavelength can be compensated using this mechanism (i.e., EO has a limited correction range). TO tuning is preferred to compensate for large shifts in MR's resonant wavelength. However, one has to compromise on latency ($\sim \mu s$ range) and power consumption, which is higher than for EO tuning. To reduce our reliance on TO tuning, which entails high overheads, we explore the possibility of a hybrid tuning mechanism, where both TO and EO tuning are used to compensate for $\Delta\lambda_{MR}$. Such a tuning method has been proposed earlier [28] and can be easily transferred to an optimized MR for hybrid tuning in our architecture. Such a mechanism would significantly reduce the overhead caused just by TO tuning.

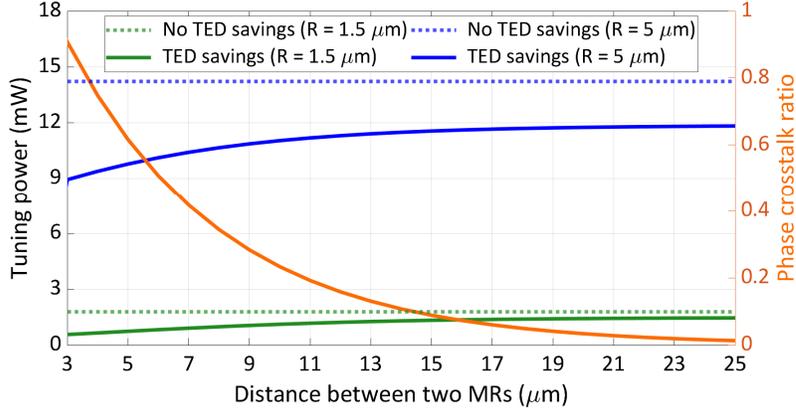

Figure 3: Tuning power compensation in a block of 10 MRs placed with and without considering thermal eigenmode decomposition (TED) for different MR radius. The orange line represents phase crosstalk ratio variation with distance between MRs.

To reduce the power overhead of TO tuning in such a hybrid approach, we adapt a method called thermal eigenmode decomposition (TED), which was first proposed in [29] that involves collectively tuning all the MRs in an MR bank. Such a tuning method proves to be more beneficial than individually tuning MRs. By doing so we can cancel the effect of crosstalk (i.e., undesired phase shift) in MRs with much lower power consumption. The amount of phase crosstalk induced from one MR on another MR, placed adjacent to each other, can be modelled using the trend in Figure 3 (orange line). In this figure, as the distance between two devices (MRs) increases, the amount of phase crosstalk between them reduces. Correspondingly, as an example we calculate the tuning power compensation for an MR bank consisting of 10 MRs and different radii placed at a distance ($d$) from each other. A few important trends to observe from Figure 3 are (i) as the radius of an MR increases, tuning power compensation for $\Delta\lambda_{MR}$ increases;(ii) Without TED (collective tuning of MRs), the tuning power consumption is high, indicating that each MR would require more power to compensate for respective shifts in resonant wavelength ($\Delta\lambda_{MR}$); (iii) By employing TED, we see a significant reduction in tuning power consumption: 51% (radius of 1.5 $\mu m$) and 41% (radius of 5 $\mu m$) when MRs are placed at a distance of 5 $\mu m$ and 7 $\mu m$ apart from each other, respectively. Though placing MRs further close to each other would yield better compensation in power, one must take into account the placement and routing of tuning circuit for each MR in an MR bank. Additional power reduction can be obtained by performing device level optimizations. Designing MRs tolerant to FPVs would reduce the total power used to compensate for fabrication variations.

**5.2 Device-level optimization**

We explore different MR designs to accommodate different needs in our *ROBIN* architecture such as multi-bit precision for activation values, single-bit precision for weight value representation, and batch normalization.

*5.2.1* Fabrication Process Variation Resilience

FPVs cause undesirable changes in device critical dimensions (e.g., width and thickness), which cause resonant wavelength shifts ($\Delta\lambda_{MR}$). To address $\Delta\lambda_{MR}$, we explore the impact of change in device parameters such as waveguide width, thickness, gap between input and ring waveguide, and radius using our in-house MR device-exploration tool. We map the behavior of different changes in the waveguide width, thickness, and radius in MRs due to FPVs. Figure 4(a) shows one of our design exploration results where we understand and observe the behavior of resonant resonant-wavelength shift slopes due to change variations in the waveguide width, thickness and radius represented by orange, green and blue lines respectively. Resonant wavelength shift slope due to change in waveguide width ($\partial \lambda_{MR}/\partial w$) can be given as :

$$\frac{\partial \lambda_{MR}}{\partial w} = \left| \frac{\Delta\lambda_{MR}(\lambda, w+\epsilon_w, t, R) - \Delta\lambda_{MR}(\lambda, w-\epsilon_w, t, R)}{2\epsilon_w} \right|. \quad (1)$$

In (1), $\epsilon_w$ denotes a small change in waveguide width and $\Delta\lambda_{MR}$ depends on changes in width ($w$), thickness ($t$), and radius ($R$). Similarly, $\partial\lambda_{MR}/\partial(t, R)$ can also be approximated.

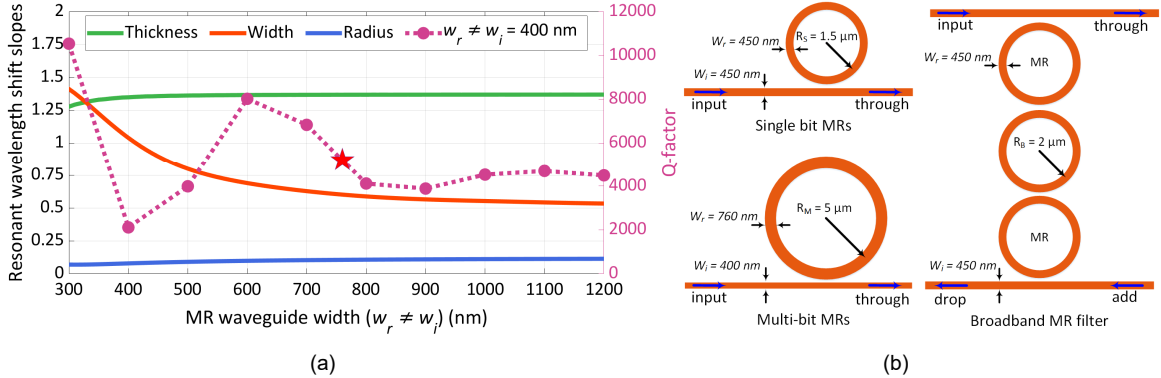

Figure 4: (a) Resonant-wavelength shift slopes with respect to changes in waveguide width, thickness, and radius, and corresponding cross-over coupling($\kappa$), when the input waveguide ($w_i$) is set to 400 nm the marked point represents our selected MR design; (b) The different MR designs considered in this work.

Figure 4(a) clearly shows that the impact of resonant-wavelength shift reduces as we increase the waveguide width, whereas the impact of thickness and radius variations remains constant. $\Delta\lambda_{MR}$ is more sensitive to changes in waveguide width, hence the impact of $\Delta\lambda_{MR}$ reduces we increase the waveguide width. We employ Lumerical MODE [30], an Eigen mode solver to calculate these shifts in resonant wavelengths. One can easily overcome higher-order mode excitation by employing adiabatic designs [30] and waveguide tapers [31] in MRs with wider waveguides. Such a design translates to lesser tuning-power consumption due to FPVs.

### 5.2.2 Multi-Bit Precision MRs

As discussed in section 4, increasing the number of bits used to capture activations in a model can boost the model accuracy in our architecture. However, there is not a significant accuracy boost beyond 4-bit activation values, hence we explore MR designs which can achieve a resolution of 4 bits. To achieve a resolution of 4 bits, we observe how the optical signals from MRs impact each other due to crosstalk. We consider calculations from [32] to define the amount of noise from one MR on the other:

$$\phi(i,j) = \frac{\delta^2}{(\lambda_i - \lambda_j)^2 + \delta^2} \quad (2)$$

where, $\phi(i-j)$ describes the noise content from the $j^{th}$ MR present in the signal from the $i^{th}$ MR, $(\lambda_i - \lambda_j)$ is the difference between the resonant wavelengths ($\lambda_i, \lambda_j$), and $\delta = \lambda/(2 \cdot Q - factor)$. Where, Quality factor or Q-factor is a measure of the sharpness of the resonance relative to the central frequency of a microring resonators (MR) that impacts the optical channel spacing, crosstalk, bandwidth, and other factors in the MR [33]. A sharper resonance (i.e., a higher Q-factor) can result in increased susceptibility to noise, as even a small change in the central frequency of the MR (due to perturbance) can lead to large losses. This effectively limits the achievable resolution of the parameters being represented. Thus, smaller Q-factors are preferred. However, a smaller Q-factor can also lead to larger device dimensions and higher optical crosstalk, which in turn can lead to larger losses and higher tuning power requirements. Q-factor in an MR is defined as follows:

$$Q\text{-}factor = \frac{\lambda_{MR}}{FWHM}, \quad (3)$$

where, FWHM is the full width at half maximum of a resonance spectrum which can be defined for an all-pass ring resonator (see Figure 4) as follows:

$$FWHM = \frac{(1-ra)\lambda_{MR}^2}{\pi n_g L \sqrt{ra}}, \quad (4)$$

where, $r$ is the self-coupling coefficient and $\kappa$ is the cross-over coupling coefficient. Also, $a$ is the single-amplitude transmission, including both the propagation loss in the ring and the loss in the couplers, which can be written as $a = e^{-\alpha L}$, where $\alpha$ is power attenuation coefficient. $L$ is round trip length or the circumference of the MR. In this paper, we assume a lossless coupler in our designed MRs, hence $|\kappa|^2 + |r|^2 = 1$. For ideal cases with zero attenuation, $a \approx 1$. Based on the above equations, the noise power component can thus be calculated as:

$$P_{noise} = \Sigma_i^{n-1} \phi(i,j) P_{in}[i] \quad (5)$$

For power intensity ($P_{in}$) of 1, the resolution can then be computed as:

$$Resolution = \frac{1}{\max|P_{noise}|} \quad (6)$$

To achieve a bit resolution of at least 4-bits, we need MRs with a Q-factor of ≈5000 (from (6)) while being tolerant to FPVs. Q-factor is highly sensitive to losses and change in dimensions of MR. Selecting input waveguide width of 400 nm and ring waveguide width of 760 nm, and radius (RM) of 5 $\mu m$ as shown in Figure 4(a) (blue line), provides improved tolerance to FPV, desirable Q-factor, and smaller area consumption. Such an MR design with Q-factor of 5000, allows enough levels of distinction between bits by slightly changing intensity, helps easily detect optical signal at the output port satisfying the requirement for multi-bit precision of activation values.

### 5.2.3 Single-bit MRs

In our architecture, we represent weight values with a single bit, and this requires just two levels of precision with an MR. An MR of high Q factor may be used here, as we don't have to have high resolution here. Compact ring designs with high Q-factor have been proposed in [34], [35]. The work in [35] proposes an MR design with radius 1.5 $\mu m$ to achieve a high Q-factor of 46,000 without the consideration of sidewall roughness while maintaining low bending loss ≈7 cm[-1]. Similarly, an adiabatic MR structure of radius 3 $\mu m$ is designed in [34] to avoid higher order mode excitation where a high Q-factor of 27,000 is achieved. These works prove our point that such high Q-factor rings can be designed.

We design a ring of radius 1.5 $\mu m$, as shown in Figure 4(b), with input waveguide ($w_i$) and ring waveguide ($w_r$) width set to 450 nm, to achieve a Q-factor of 25,000 that corresponds to a bit resolution of 1 from (6). These designs allow our architecture to save on area and tuning power consumption (Figure 3). We acknowledge that FPVs are an inevitable part of the fabrication process. However, since we just need to differentiate between two levels of operations, we do not explore for designs that are tolerant towards FPVs, for single-bit MRs.

*5.2.4 Broadband MRs*

Batch normalization (BN) layers can be considered essential in BNNs as they add complexity to the models, via the gain ($\gamma$) and bias ($\beta$) terms of the layer. These terms are learned during the training process along with the normalization parameters of the batch mean ($\mu$) and standard deviation ($\sigma$). As they are being learned in training, these terms are dynamic, but during inference they have static values. This allows us to have a photonic version of batch normalization folding, where we may tune weights as per the following equation:

$$w_{fold} = \gamma \cdot \frac{W}{\sqrt{\sigma^2 + \epsilon}} = C_{fold} \cdot W. \tag{7}$$

There is a similar equation for bias terms as well, but since BNN models benefit from batch normalization after every layer, these will be normalized out and hence can be ignored. The above constant, $C_{fold}$ is applied to every weight term and hence is a participant in every matrix multiplication operation, i.e.:

$$Input_{l+1} = f\left(A_l \cdot (w_{fold})_l\right) = C_{fold} f(A_l \cdot W_l). \tag{8}$$

In (8), $Input_{l+1}$ refers to the input to the $(l+1)$[th] layer, $f()$ is non-linear activation function, $A_l$ is the activation of $l$[th] layer and $W_l$ is the weights from $l$[th] layer. This operation can be applied to partial sums as well, and can be implemented using a broadband photonic device with its gain tuned to reflect $C_{fold}$. The broadband device is preferred as this allows simultaneous gain tuning of all the wavelengths in the waveguide efficiently, both area and energy wise.

Hence, the last type of MRs we consider are broadband MRs that are needed for batch normalization (BN) layers due to their relevance in BNNs. A large passband can be achieved by cascading several MRs and properly selecting the design parameters of MRs [36]. We explore such a higher order MR, or cascaded MR filter, to achieve a wide passband. [37] explores a possibility for passband widths ranging from 6.25 Ghz to a maximum of 3 Thz. This work explores different design parameters of a higher-order filter while evaluating different losses such as insertion, propagation and coupling loss in higher order MRs. A 0.5 nm resonant wavelength shift of MR was reported for a fabrication error of 10 nm showing that such a design is tolerant to FPVs.

A 3rd-order MR based switching device with radius of 2 $\mu m$ shown in Figure 4(b), fits the requirement for broadband MR. The coupling coefficients at the input ($\kappa_i^2$) is 0.53 and coupling at higher order rings is 0.2. The propagation loss of 25 dB/cm has been reported and insertion loss of the two elements in higher order filter are 4.35 dB and 0.36 dB, respectively [36]. Having such a design, one can achieve a flat-top passband with bandwidth width of at least 3 THz. Employing such a broadband MR can help us apply the non-linear activation on all the available resonant wavelengths in the bank. Having a large bandwidth such as 2.5 Thz allows us to conveniently tune up to 20 different wavelengths.

## 5.3 Architecture design

An overview of the *ROBIN* accelerator architecture is shown in Figure 5. The optical device and tuning circuit optimizations from the previous subsections are utilized within the optical binary vector dot product (VDP) units. We use banks of heterogeneous MRs to imprint activations, weights, and the BN layer constants onto optical signals. Multiple such VDP units are composed together to form the overall architecture, as shown in the figure, which is then used to accelerate a given BNN model. We utilize a photonic summation unit for summing the partial sum outputs from our VDPs, before passing the partial sums on to the electronic control unit (ECU), as shown in Figure 5. We also rely on the ECU for fetching parameters from the global memory, decomposing them to lower dimensional vectors, distributing these vectors among the VDP units, and for implementing non-linear activations functions and pooling layers. We describe the working of the *ROBIN* architecture in more detail in the following subsections.

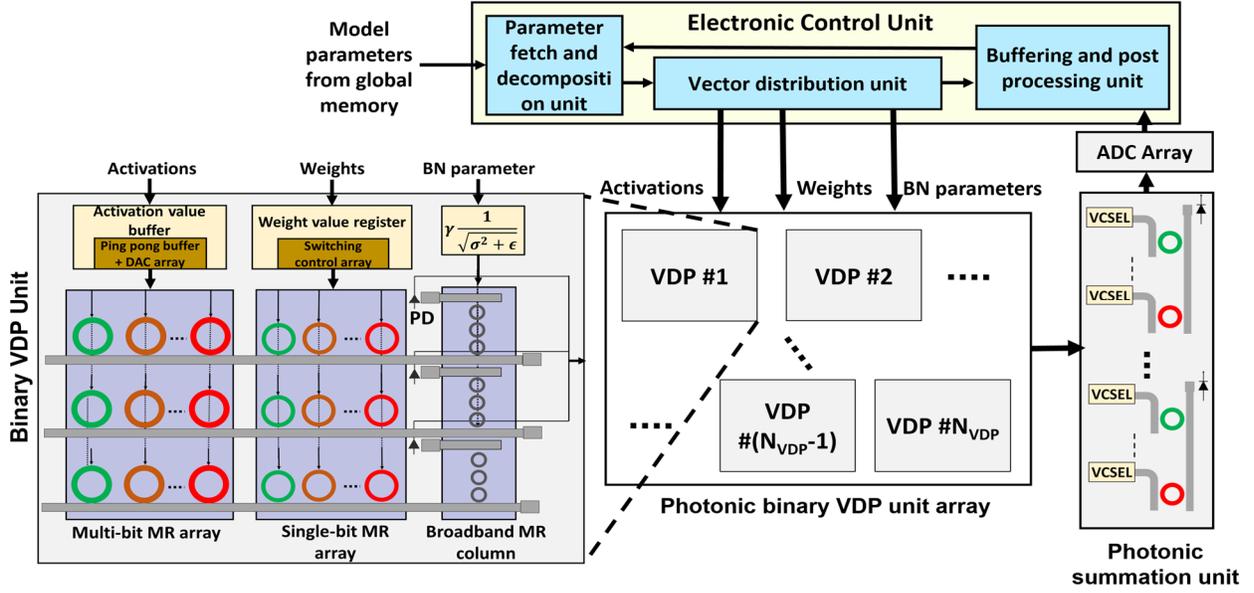

Figure 5: An overview of the *ROBIN* architecture, showing the electronic control unit, the photonic vector dot product (VDP) unit array, and the photonic summation unit, along with a detailed view of the VDP unit internal structure.

*5.3.1 Decomposing vector operations*

To map convolution (CONV) and fully connected (FC) layers from BNN models to our accelerator, we first need to decompose large vector sizes into smaller ones, so they can mapped to the VDP array in our architecture.

In CONV layers, a filter performs convolution on a patch (e.g., 2×2 elements) of the activation matrix in a channel to generate an element of the output matrix. The operation can be represented as:

$$K \otimes A = Y, \qquad (9)$$

Assuming a 2×2 filter kernel and weight matrices, (9) can be rewritten as:

$$\begin{bmatrix} k_1 & k_2 \\ k_3 & k_4 \end{bmatrix} \otimes \begin{bmatrix} a_1 & a_2 \\ a_3 & a_4 \end{bmatrix} = k_1 a_1 + k_2 a_2 + k_3 a_3 + k_4 a_4, \qquad (10)$$

Rewriting (10) as a vector dot product, we have:

$$\begin{bmatrix} k_1 & k_2 & k_3 & k_4 \end{bmatrix} \cdot \begin{bmatrix} a_1 \\ a_2 \\ a_3 \\ a_4 \end{bmatrix} = k_1 a_1 + k_2 a_2 + k_3 a_3 + k_4 a_4, \qquad (11)$$

Once we are able to represent the operation as a vector dot product, it is easy to see how it can be decomposed into partial sums. For example:

$$[k_1 \quad k_2] \cdot \begin{bmatrix} a_1 \\ a_2 \end{bmatrix} = k_1 a_1 + k_2 a_2 = PS_1, \tag{12a}$$

$$[k_3 \quad k_4] \cdot \begin{bmatrix} a_3 \\ a_4 \end{bmatrix} = k_3 a_3 + k_4 a_4 = PS_2, \tag{12b}$$

$$PS_1 + PS_2 = Y, \tag{12c}$$

In FC layers, typically much larger dimension vector multiplication operations are performed between input activations and weight matrices. Therefore, we have:

$$A \cdot W = \begin{bmatrix} a_1 \\ a_2 \\ \vdots \\ a_n \end{bmatrix} \cdot [w_1 \quad w_2 \quad \cdots \quad w_n], \tag{13}$$

$$A \cdot W = \begin{bmatrix} a_1 \cdot w_1 + a_1 \cdot w_2 + \cdots + a_1 \cdot w_n \\ a_2 \cdot w_1 + a_2 \cdot w_2 + \cdots + a_2 \cdot w_n \\ \vdots \\ a_n \cdot w_1 + a_n \cdot w_2 + \cdots + a_n \cdot w_n \end{bmatrix}, \tag{14}$$

In (13), a1 to an represent column vectors of activations (A) and w1 to wn represent row vectors of weight matrix (W). The resulting vector is a summation of dot products of vector elements (14). Similar to the decomposition of CONV operation, these can then be decomposed into lower dimensional dot products.

### 5.3.2  Vector dot product (VDP) unit design

As discussed in subsection 5.3.1, we consider matrix operations to be decomposed to lower dimensional vector dot products. These vector dot product operations are executed optically within our VDP units. The heterogeneous MR designs combined with optical circuit-level optimizations for area and power consumption are utilized to design VDP units (Figure 5) suited for accelerating both CONV and FC layers without compromising on accelerator throughput. For representing weight values, we use high Q-factor, small radius single-bit MRs described in Section 5.2.2. The smaller radius contributes to lower tuning power and reduces optical losses along the waveguide. This is possible due to the binarized nature of weight matrices in BNNs. For activation values we consider MRs with slightly lower Q-factor, for better resolution, as discussed in Section 5.2.1. For the BN layer we have to simultaneously tune all the wavelengths in the waveguide to the batch normalization constant, and for this we use third order MR filters, as described in Section 5.2.3. The combination of these heterogeneous designs allows the VDP units to be highly energy efficient. The VDP units also perform BN layer folding, which can be done efficiently in the optical domain, as discussed in Section 5.2.3. We also make use of electronic buffering in the VDP units to reduce the digital to analog converter (DAC) usage. In particular, we make use of ping-pong buffers, which allow us to use a single DAC array to feed the activation devices in all the waveguides in a VDP unit. As weight values are single-bit values, we can use simple switching circuits to essentially turn the MRs on or off depending on the value of the weight parameters.

In designing a VDP unit, there are several important parameters that must be carefully considered: number of higher resolution MRs for activation representation ($N_A$), number of single-bit MRs for weight representation ($N_W$), and number of broadband MRs ($N_B$) for batch normalization folding implementation. Thus, the total number of MRs per waveguide NMR = $N_A + N_W + N_B$. The number of required DACs is equal to NA. The number of waveguides to which we distribute the MRs is denoted as $N_{WG}$. The size of the vector represented in the VDP unit is given by NWG*NA. We divide this vector across multiple waveguides to reduce power consumption, as this allows us to reuse wavelengths and reduce the overall laser power consumption, as discussed next, in subsection 5.3.3. For efficient parallelization and to increase the throughput of the accelerator, multiple VDP units work concurrently on parameters from the same layer and generate partial sums simultaneously. The total VDP unit count used in ROBIN is NVDP. Thus the VDP and architecture design process can be considered as an optimization problem where we try to explore NVDP, NWG, NA (= NW), and NB values while trying to maximize throughput and minimize area and power consumption. We present results of this architecture exploration analysis in Section 6.3.

### 5.3.3 Optical wavelength reuse in VDP units

Prior works on optical accelerator design typically considers a separate wavelength to represent each individual element of a vector. This approach leads to an increase in the total number of lasers needed in the laser bank (as the size of the vectors increases) which in turn increases power consumption. Beyond employing the decomposition approach discussed above, we also consider wavelength reuse per VDP unit to minimize laser power. In this approach, within VDP units, the vectors assigned from the electronic control unit (ECU) are further decomposed into smaller sized vectors for which dot products can be performed using MRs in parallel, in each arm of the VDP unit. The same wavelengths can then be reused across arms within a VDP to reduce the number of unique wavelengths required from the laser. Photodetectors (PDs) perform summation of the element-wise products to generate partial sums from decomposed vector dot products. The partial sums from the decomposed operations are then converted back to the optical domain by VCSELs (bottom right of Figure 5), multiplexed into a single waveguide, and accumulated using another PD, before being sent for buffering. Thus, our approach leads to an increase in the number of PDs compared to other accelerators but significantly reduces both the number of MRs per waveguide and the overall laser power consumption. The reduction in overall power consumption is also assisted by the fact that PDs do not consume significant power.

In each arm within a VDP unit, we can use a maximum of 15 MRs per bank for a total of 30 MRs per arm. The choice of MRs per arm considers not only the thermal crosstalk and layout spacing issues and the benefits of wavelength reuse (as discussed earlier), but also the fact that optical splitter losses become non-negligible as the number of MRs per arm increase, which in turn increases laser power requirements. Thus, the selection of MRs per arm within a VDP unit must be carefully adjusted to balance parallelism within/across arms, and laser power overheads.

### 5.3.4 ROBIN pipelining and scheduling

The pipeline and schedule of operations during BNN model execution on the *ROBIN* accelerator is shown in Figure 6. The electronic control unit (ECU) for the accelerator communicates with the global memory and retrieves the trained weights for the model being accelerated. The weights are stored in SRAM-based buffers. Considering the vector granularity of the VDP units, latency of operation of the photonic core, and the parameter sizes (4-bit activation bits and binary weight parameters), we can calculate the memory bandwidth necessary. From our analyses (presented in Section 6.3), we found that our architecture needs a maximum bandwidth of 93.75 GB/s at the ECU to photonic core interface. This is a reasonable bandwidth assumption for an SRAM-based memory with operating frequency ≥ 2.5 GHz and a read width of 250 bits. Previous works, such as [39], have explored similar SRAM systems, but for a much higher bandwidth requirement at 250 GB/s. The lower bandwidth requirement for our system can be attributed to the smaller parameter sizes, while the work in [39] considered 16-bit precision for the neural network parameters. Memory interfaces which exceed the bandwidth necessary are already available commercially: e.g., NVIDIA Tesla K20M GPUs have 320-bit memory interfaces at 2.6 GHz which can operate every half clock cycle to provide a bandwidth of 208 GB/s.

These weight matrices are decomposed to lower dimensional vectors and are distributed to the VDPs by the ECU's vector decomposition unit. The decomposition operation is described by the left-hand side of equations 10 to 12. As described in the equations, the vector decomposition unit converts matrices to vectors (row-wise conversion for weight matrices and column-wise conversion for column matrices), and then those vectors into sub vectors. The size of the sub vectors depends on the granularity of the VDP units. The received vectors are buffered in the VDP units and are fed into the DAC array through a ping-pong buffer so that they can keep the MAC operation running continuously. The partial sums generated are passed on to the photonic summation unit, the output from which is passed on to the ECU. The ECU buffers the sums and calculates inputs that are then passed on to the next layer by subjecting the parameters to non-linearities (activation functions) and performing other layer specific operations, like pooling.

The model parameter buffering stage is not repeated every pipeline operation, but must be repeated as the parameters buffered in the buffers in ECU are depleted (i.e., distributed to VDP units). As such, the total time required by *ROBIN* to perform inference acceleration for a given model can be given as:

$$Total\ time\ of\ operation = T_{del} + \Delta t \times X + (ECU\ parameter\ buffering\ delay) \times x, \qquad (15)$$

where,

$$\Delta t = local\ buffer\ operation\ delay + vector\ distribution\ delay, \qquad (16)$$

$$X = \frac{Total\ number\ of\ parameters\ in\ the\ model}{N_w \times N_{VDP}}, \qquad (17)$$

$$x = \frac{(Parameters\ buffered\ in\ ECU)}{N_w \times N_{VDP}}, \qquad (18)$$

Comparing our pipeline to the pipeline presented in the previous work on photonic BNN acceleration, [21], we can observe the following differences: (i) ROBIN's pipeline takes into consideration model parameter retrieval from global memory, buffering in the ECU, and how these parameters are utilized in the photonic core. The pipeline in [21] does not include these operations in its pipeline.; (ii) ROBIN's pipeline considers both ECU and photonic core operation, whereas the pipeline in [21] is photonic system centric; (iii) ROBIN utilizes photonic batch normalization folding which does not require an extra step, whereas in [21] this operation is performed electronically and requires a separate stage in their pipeline.

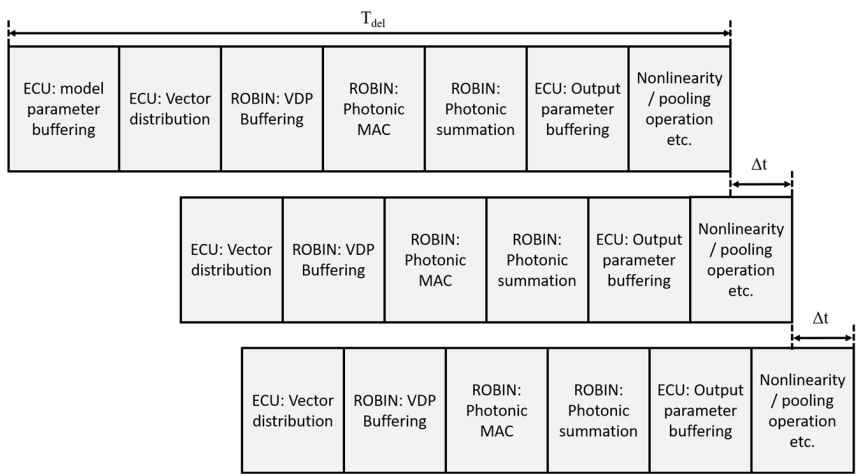

Figure 6: Pipelined scheduling of operations during BNN execution on the *ROBIN* accelerator

Table 1: Models and datasets used for evaluations

| Model no. | CONV layers | FC Layers | BN layers | Parameters | Datasets |
|---|---|---|---|---|---|
| 1 | 2 | 2 | 3 | 60,642 | Sign MNIST |
| 2 | 6 | 3 | 6 | 1,546,570 | CIFAR10 |
| 3 | 6 | 3 | 7 | 13,570,186 | STL10 |
| 4 | 6 | 2 | 6 | 552,362 | SVHN |

## 6 RESULTS

### 6.1 Simulation setup

We conducted several simulation studies to evaluate the effectiveness of our BNN accelerator. The optimized heterogeneous MR designs, the tuning circuit optimizations, and architectural level considerations discussed so far were included in our simulation considerations. For these simulations, we considered four binarized DNN models shown in Table 1. Model 1 is LeNet5 [41], binarized and with BN layers added. The other three models are custom BNNs. The datasets used to train these models are also provided in Table 1. We implemented and simulated the *ROBIN* architecture using a custom Python simulator to estimate its performance in terms of power, frames per second (FPS) performance, and energy consumption. For analyzing the inference accuracy across different activation precision and the impact of FPV noise on the inference accuracy, we used Tensorflow 2.3 along with Qkeras [42]. Figure 7 shows the training accuracy versus epoch graph of the models described in Table 1, to illustrate the accuracy and loss across the epochs.

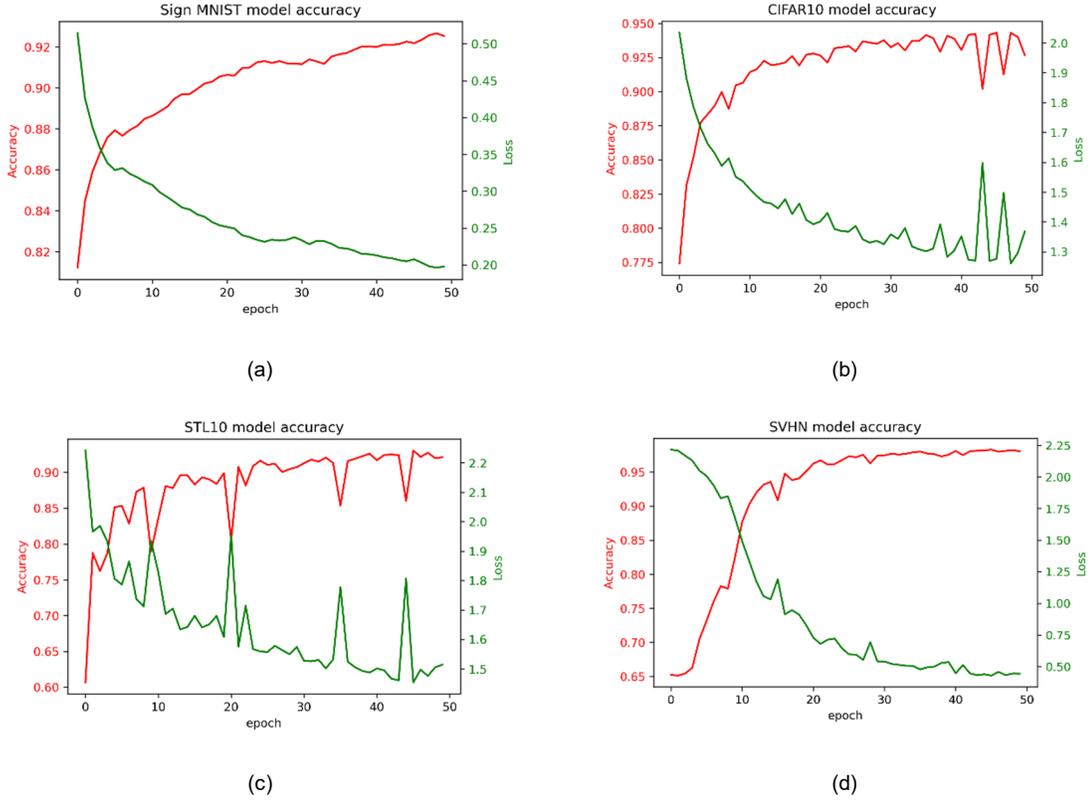

Figure 7: The training accuracy vs epoch for the BNN models considered for (a) Sign MNIST, (b) CIFAR10, (c) STL10, and (d) SVHN datasets. (a) shows top-1 accuracy, while (b)-(d) show top-5 accuracy.

Table 2: Parameters considered for analysis of photonic accelerators

| Devices | Latency | Power |
|---|---|---|
| EO Tuning [24] | 20 ns | 4 $\mu W$/nm |
| TO Tuning [11] | 4 $\mu s$ | 27.5 mW/FSR |
| VCSEL [44] | 10 ns | 0.66 mW |
| TIA [45] | 0.15 ns | 7.2 mW |
| Photodetector [46] | 5.8 ps | 2.8 mW |
| DAC [47] | 0.33 ns | 59.7 mW |
| ADC [48] | 24 ns | 62 mW |

We compare *ROBIN* with DEAP-CNN [14] and HolyLight [15], two recent optical DNN accelerators from prior work, along with LightBulb [21], which is an optical BNN accelerator, as well as numbers reported from several electronic DNN and BNN accelerators. For the optical accelerators, we considered optical signal losses due to various factors: signal propagation loss (1 dB/cm [8]), splitter loss (0.13 dB [49]), combiner loss (0.9 dB [50]), MR through loss (0.02 dB [51]), MR modulation loss (0.72 dB [52]), microdisk loss (1.22 dB [53]), EO tuning loss (6 dB/cm [24]), and TO tuning loss (1 dB/cm [11]). We also considered the ADC design from [48] and the 4-bit DAC from [47] in our analyses. The analysis of the optical accelerators (DEAP-CNN [14], HolyLight [15], and LightBulb [21]) follows the modeling methodology we have adopted for ROBIN, where we factor in power consumption and delays associated with photonic devices used in these accelerators. A summary of the power and latency considerations for our analyses is given in Table 2. These power and latency values were used in our simulations and latency of operation of our architecture. A comparison for inference time on ROBIN and a conventional CPU is presented in Section 6.5.

To calculate laser power consumption, we use the following power model:

$$P_{laser} - S_{detector} \geq P_{photo-loss} + 10 \times \log_{10} N_\lambda, \tag{19}$$

where $P_{laser}$ is the laser power in dBm, $S_{detector}$ is the PD sensitivity in dBm, and $P_{photo-loss}$ is the total loss encountered by the optical signal, due to all of the factors discussed above.

### 6.2 Fabrication-process variation analysis

FPV in optical devices is corrected using TED tuning in our architecture, as discussed in Section 5. At the system level, this tuning leads to significant power consumption overhead, and any avenue to further reduce tuning power consumption becomes important. To this end we conduct an FPV noise injection analysis, where we inject noise, modeled using FPV data, into the MR devices into our *ROBIN* accelerator, during the inference phase. This experiment was conducted to *(i)* study the impact of FPV induced noise on BNN models mapped to our accelerator, *(ii)* determine how effective TED tuning is in such scenarios, and *(iii)* uncover any opportunities for further power minimization.

To analyze the impact of FPV on the model and how TED tuning compensates for it, we first consider the effect of FPV on the shift in resonant wavelength ($\Delta\lambda_{MR}$) in MRs. Resonant-wavelength shift in an MR can be modeled from [38] as:

$$\Delta\lambda_{MR} = \frac{\partial\lambda_{MR}}{\partial w}\sigma_w + \frac{\partial\lambda_{MR}}{\partial t}\sigma_t + \frac{\partial\lambda_{MR}}{\partial R}\sigma_R, \tag{20}$$

where, $\sigma_{w,t,R}$ are the associated standard deviations for waveguide width, thickness, and radius variations; and $\frac{\partial\lambda_{MR}}{\partial(w,t,r)}$ is the rate of change in the MR resonant wavelength considering the variations in the waveguide width, thickness, and radius represented in (20). We generate virtual FPV maps with a mean ($\mu$) of 0 and standard deviation ($\sigma_{(w,t,R)}$) of 4.9 nm, 1.5 nm, and 0.75 nm for waveguide width, thickness, and radius, respectively. These standard deviation values are experimentally obtained based on real fabricated MR devices through our collaboration with CEA-Leti. Using these values, we are able to derive $\Delta\lambda_{MR}$ using (20). So, the current resonant wavelength ($\lambda'_{MR}$) of the FPV affected MR becomes:

$$\lambda'_{MR} = \lambda_{MR} + \Delta\lambda_{MR}, \tag{21}$$

Due to a shift in $\lambda_{MR}$, the transmission of the wavelength through the MR is impacted. The intensity of the wavelength at the through port is given by the following equation from [33].

$$T = \frac{I_{out}}{I_{in}} = \frac{a^2 - 2ra\cos\phi + r^2}{1 - 2ar\cos\phi + ra^2}, \tag{22}$$

In (22), $\phi = \beta L$, with $L$ being the roundtrip length and $\beta$ the propagation constant $\beta = 2\pi/\lambda$ of the circulating mode; and $r^2$ is the self-coupling coefficient of an MR. A detailed analyses for the calculation of $r$ using super mode theory is presented in [40]. The output intensity from the MR is important, as for noncoherent MAC units, the parameter values are encoded onto the signal intensity, and a change in expected output can be seen as perturbation or noise source.

The noise injection was modeled using equations (20) and (22), where we consider the resonant-wavelength shift ($\Delta\lambda_{MR}$) in MRs due to FPV and its impact on the parameters imprinted on the MRs. From our analysis using the FPV data from our device fabrications with CEA-Leti, and equation (18), we are able to obtain the mean and standard deviation values for $\Delta\lambda_{MR}$ in a wafer. The values calculated are $\mu = $ -0.1461 nm and $\sigma = $ 24.417 nm. Using these values, 50 $\Delta\lambda_{MR}$ maps for the accelerator were generated and then using equation (22) the perturbation to the parameters imprinted on to the devices were modeled. Noise injection to the models was performed at inference time using Tensorflow.

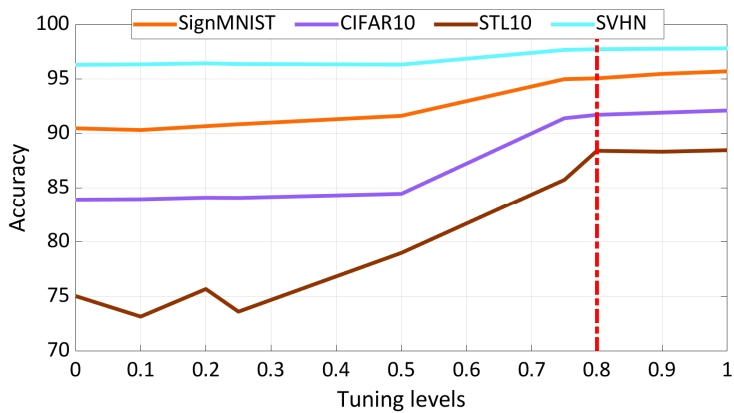

Figure 8: Inference accuracy versus level of tuning applied. At 80% tuning, the inference accuracy saturates, rendering further tuning unnecessary, and providing an opportunity to save tuning power.

Figure 8 shows the results of this experiment, where we explored the impact of FPV-induced noise in the four BNN models, and the effect of TED tuning for FPV compensation. We expected that the better the devices were tuned, the better the accuracy that would be exhibited by the accelerator. But it was observed that the model's accuracy can be sustained without completely (perfectly) tuning the devices. Figure 8 shows that at 80% FPV correction through tuning, the BNN retains appreciable inference accuracy. Thus there is not a significant accuracy benefit to tune beyond the 80% level, and staying with the 80% level can allow us to achieve power savings. This reduction in tuning power is factored into our architecture level analysis, which is presented next.

### 6.3 ROBIN architecture optimization analysis

In this section, we show results of our exploration of the parameters discussed in Section 5.3.2. As mentioned in Section 5.3.2, we try to optimize $N_{VDP}$, $N_{WG}$, $N_A$, and $N_B$ to reduce area and power consumption while trying to obtain the best throughput (frames per second or FPS) possible. $N_B$ was fixed to be 1 per waveguide, allowing us to have up to 20 wavelengths in the same waveguide with a channel spacing of 1 nm, which in turn allows us to tune all of the MRs simultaneously to the BN layer parameters. We then explored $N_{VDP}$, $N_{WG}$, and $N_A$, with the goal of optimizing power, area, and FPS. The result of this exploration analysis is shown in Figure 9 in the form of a scatter plot. From this analysis we identified two configurations for *ROBIN*, where one is optimized for FPS/Watt, with lowest area and power consumption (energy optimized *ROBIN* or *ROBIN-EO*), and another with the best FPS but with higher area and power consumption (performance optimized *ROBIN* or *ROBIN-PO*). In terms of ($N_A, N_{VDP}, N_{WG}$), these configurations can be represented as (10, 50, 10) for *ROBIN-EO* and (50, 200, 10) for *ROBIN-PO*. To explore the efficiency of these configurations, they are compared against other optical and electronic DNN/BNN accelerator platforms. Results for these comparisons with other accelerators are presented in the following section.

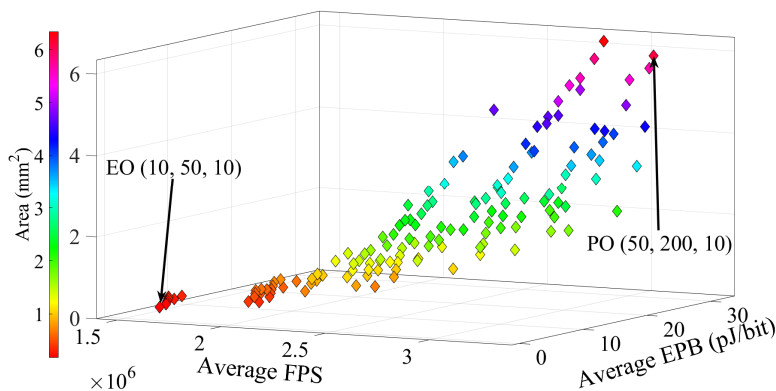

Figure 9: Scatterplot of average FPS vs. average EPB vs. area of various *ROBIN* configurations. The configuration with highest FPS/Watt (energy optimized or EO) and the one with best FPS (performance optimized or PO) are specified.

### 6.4 Comparison with state-of-the-art optical and electronic DNN/BNN accelerators

We compared *ROBIN-EO* and *ROBIN-PO* against various electronic and optical neural network acceleration platforms. For optical DNN accelerator platforms, we selected DEAP-CNN [14] and HolyLight [15]. For electronic DNN accelerator platforms, we compare against a GPU (Nvidia Tesla P100), along with several architectures including SIGMA [54], Edge TPU [55], DaDianNao [56], and FPGA implementation of Null Hop [57]. We also compare *ROBIN* against the best known previous work on photonic BNN accelerators, LightBulb [21]. We also compare *ROBIN* against the best-known previous work on photonic BNN accelerators, LightBulb [21]. When compared to LightBulb, *ROBIN* has the following differences: (i) *ROBIN* is designed to accelerate partially binarized neural networks, as opposed to fully binarized neural networks as in [21], for obtaining better accuracies; (ii) *ROBIN* utilizes photonic batch normalization folding for faster, energy efficient batch normalization layer operation whereas [21] relies on an electronic implementation of the batch normalization operation; (iii) *ROBIN* has various circuit- and device-level optimizations in place to counteract thermal and process variations, which also ensure high throughput and energy efficient operation; whereas [21] does not take into account thermal and process variations and the necessary tuning latency and energy consumption overheads needed to counter them; (iv) architecture-level optimizations in *ROBIN* ensure lower power consumption in terms of tuning and laser power; these considerations are not part of the architecture proposed in [21]. We also compare against electronic BNN accelerators FBNA [58] and FINN [59]. We used the GOPS and power consumption parameters from [60] and [8] to simulate inference on the electronic platforms.

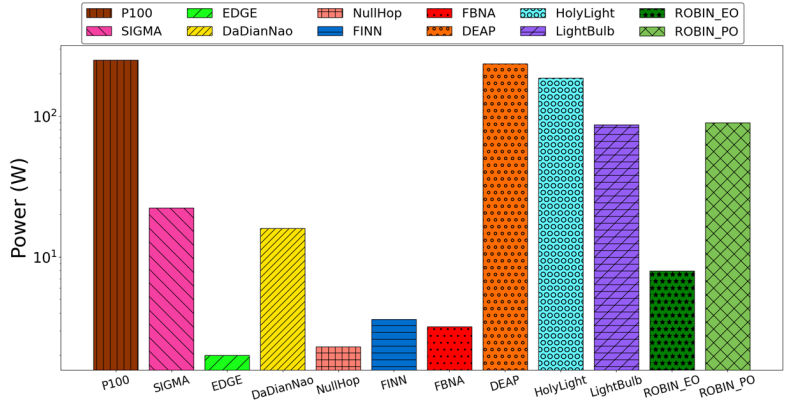

Figure 10: Power consumption comparison among variants of *ROBIN* versus other optical accelerators (DEAP-CNN, Holylight, LightBulb), and electronic accelerator platforms (P100, SIGMA, EdgeTPU, DaDianNao, Null Hop, FINN, and FBNA)

Figure 10 shows the power comparison across the accelerators from prior work and the two *ROBIN* variants. It can be observed that *ROBIN-PO* has substantially higher power consumption than *ROBIN-EO*, as *ROBIN-PO* is focused on FPS performance rather than energy conservation. *ROBIN-PO* has a much larger vector granularity per VDP unit along with substantially higher VDP unit count to maximize parallelism, when compared to *ROBIN-EO*. The larger unit count and the waveguide count in *ROBIN-PO* drives its power requirements higher. On the other hand, it can be observed that the energy and area efficient *ROBIN-EO* has comparable power consumption to that of edge and mobile electronic neural network accelerators.

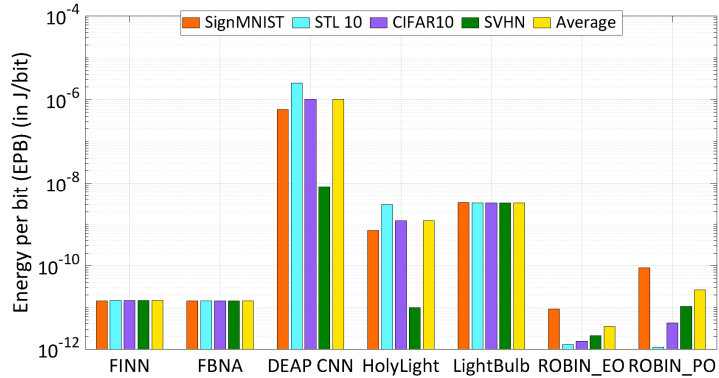

Figure 11: EPB comparison between electrical BNN accelerators, optical accelerators, and the *ROBIN* variants

In Figure 11, we compare the energy-per-bit values (EPB) across the various BNN accelerators considered in this work. We can observe that both the *ROBIN* variants perform significantly better than the optical accelerators in comparison. This lower EPB is owing to the meticulous device, circuit, and architecture level optimizations we have considered in our architecture, which takes into account various losses and delays at the architecture level and counteracts them. The heterogeneous MRs used in *ROBIN* provide energy and area benefits, and the utilization of TED for collectively tuning MRs provides further energy benefits on top of the 20% reduction we obtained from the analysis in section 6.3. TED also allows for closer placement of MRs, which in turn helps reduce propagation delays. This reduction is also impacted by the faster inputs to DAC arrays enabled by local buffering and ping-pong buffers in the VDP units.

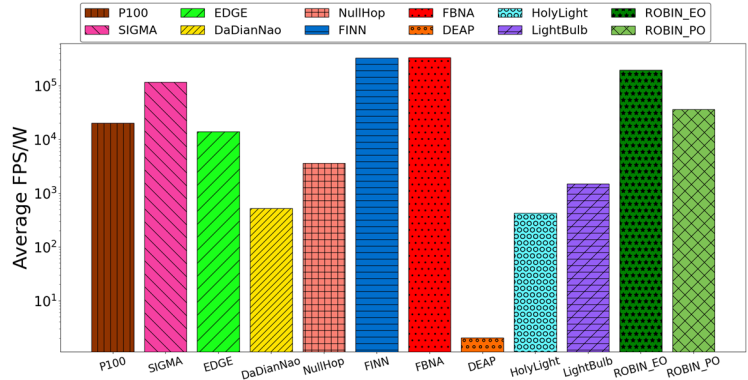

Figure 12: Average FPS/Watt among different accelerator platforms, visualized.

Lastly, in Figure 12 we present the average FPS/Watt comparison between the various accelerator platforms. Both the *ROBIN* variants perform well against the accelerator platforms to which they were compared against. *ROBIN-EO* outperforms all other platforms other than FBNA and FINN. This is owing to the extremely low power consumption reported by these BNN accelerators. However, the ROBIN variants display superior FPS performance with respect to these electronic accelerators, as can be seen in Figure 13.

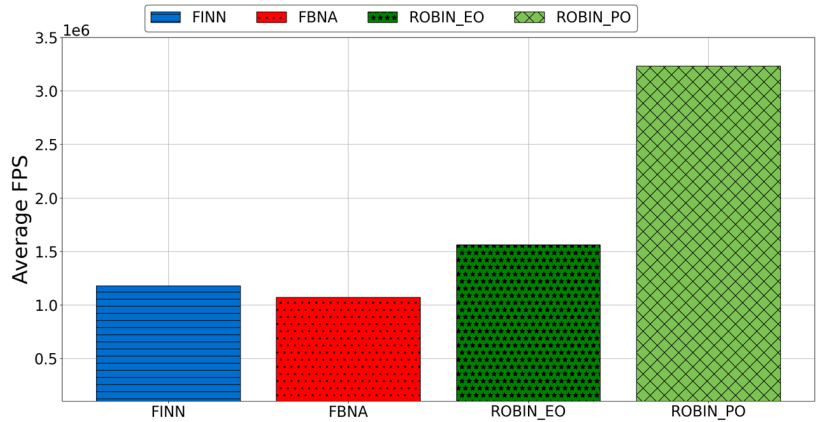

Figure 13: FPS comparison between the *ROBIN* variants and the electronic BNN accelerators.

In summary, this work showcases the effectiveness of cross-layer design of BNN accelerators with the emerging silicon photonics technology for energy/area efficient implementations and for performance-oriented designs. Overall, we can see that our energy efficient design (*ROBIN-EO*) exhibits EPB values ~4× lower than electronic BNN accelerators and ~933× lower than the photonic BNN accelerator, while the performance oriented design (*ROBIN-PO*) shows ~3x and ~25x better FPS than the electronic and photonic BNN accelerators respectively. With the growing maturity of silicon photonic device fabrication in CMOS-compatible processes, it is expected that the energy costs of device tuning, losses, and laser power overheads will go further down, making an even stronger case for considering optical-domain accelerators for deep learning inference.

Table 3: Inference time on *ROBIN-PO* and Intel i7 desktop for the four models

| Model no. | Parameters | Datasets | Inference time (for one image) | |
|---|---|---|---|---|
| | | | *ROBIN-PO* | i7-4790 |
| 1 | 60,642 | Sign MNIST | 0.0218 $\mu$s | 0.16 ms |
| 2 | 1,546,570 | CIFAR10 | 0.28 $\mu$s | 1.75 ms |
| 3 | 13,570,186 | STL10 | 2.3 $\mu$s | 2.5 ms |
| 4 | 552,362 | SVHN | 0.11 $\mu$s | 1.25 ms |

## 6.5 Comparison to CPU based inference

To highlight the advantage of dedicated inference acceleration, we have compared the performance of our ROBIN architecture against a standard desktop CPU performing inference on these models. The CPU we have considered is an Intel i7-4790, and we have used Tensorflow to analyze the latency for inference. The CPU, i7-4790, is reported to have an average power consumption of approximately 103W. This power consumption is comparable to the ~90 W we report for the ROBIN-PO variant. The summary of observations for inference time are shown in Table 3. It can be clearly observed that the ROBIN accelerator provides several orders of magnitude reduction in inference time for all the four models and datasets, compared to the Intel i7 system.

## 7 CONCLUSION AND FUTURE WORK

In this paper we proposed *ROBIN,* an optical-domain BNN accelerator which utilizes device-level, circuit-level and architecture-level optimizations to save on energy and area while improving overall throughput. Through our optimization efforts we identified two variants of *ROBIN*: *ROBIN-EO*, which is optimized for energy and area efficiency, and *ROBIN-PO*, which exhibits higher FPS performance, at the expense of greater power consumption. Our simulation analysis showed that *ROBIN* exhibits significantly better EPB performance than the various state-of-the-art optical neural network accelerators. Owing to significantly lower power consumption reported by the electronic BNN accelerators considered, *ROBIN* variants are not able to obtain better FPS/Watt than them, but upon closer examination both *ROBIN* variants can be seen to have better throughput than the electronic BNN accelerators. These results highlight the promise of our proposed *ROBIN* accelerator for accelerating BNN model execution for resource-constrained platforms.

The work described in this paper is focused on BNN acceleration using photonic systems. In this work, we considered how photonic systems can be used to accelerate the partially binarized networks, with weights remaining binary, while activations being multi-bit parameters. An immediate consideration for extension is to employ mixed quantization in the models considered, where different layers have different levels of quantization for their activation parameters. This can enable better accuracy for the considered BNN models. The photonic system- and device-level optimizations are not limited to BNN inference accelerators. These techniques may also be considered for other non-BNN accelerators for DNN/CNNs as well.

[49] L. H. Frandsen et al., "Ultralow-loss 3-dB photonic crystal waveguide splitter," in *Optics letters*, vol. 29, no. 14, 2004.

[50] Y. Tu et al., "High-Efficiency ultra-broadband multi-tip edge couplers for integration of distributed feedback laser with silicon-on-insulator waveguide," in *IEEE Photonic Journal*, vol. 11, no. 4, 2019.

[51] S. Bahirat, S. Pasricha, "OPAL: A multi-layer hybrid photonic NoC for 3D ICs," in *IEEE/ACM ASPDAC*, 2011.

[52] H. Jayatileka et al., "Crosstalk limitations of Microring-Resonator based WDM Demultiplexers on SOI," in *OIC* 2015.

[53] E. Timurdogan et al., "Vertical junction silicon microdisk modulator with integrated thermal tuner," in *CLEO:Science and Innovations*, OSA, 2013.

[54] E. Qin et al., "SIGMA: A Sparse and Irregular GEMM Accelerator with Flexible Interconnects for DNN Training," in *IEEE HPCA*, 2020.

[55] S. Cass, "Taking AI to the edge: Google's TPU now comes in a maker-friendly package," in IEEE Spectrum, vol. 56, no. 5, pp. 16-17, May 2019.

[56] T. Luo et al., "DaDianNao: A Neural Network Supercomputer," in IEEE Transactions on Computers, vol. 66, no. 1, pp. 73-88, 1 Jan. 2017.

[57] A. Aimar et al., "NullHop: A Flexible Convolutional Neural Network Accelerator Based on Sparse Representations of Feature Maps," in *IEEE Trans. Neural Netw. Learn. Syst.,* vol. 30, no.3, pp. 644-656, March 2016.

[58] P. Guo et al., "FBNA: A fully Binarized Neural Network Accelerator," *International Conference on Field Programmable Logic and Applications,* 2018.

[59] Y. Umuroglu et al.,"FINN: A Framework for Fast, Scalable Binarized Neural Network Inference," in *ACM/SIGDA FPGA,* 2017.

[60] M. Capra et al., "An updated survey of efficient hardware architectures for accelerating deep convolutional neural networks", in *Future Internet* 2020.
24